\documentclass[conference]{IEEEtran}
\IEEEoverridecommandlockouts
\usepackage{cite}
\usepackage{subcaption}
\usepackage{amsmath,amssymb,amsfonts}
\usepackage{algorithmic}
\usepackage{graphicx}
\usepackage{textcomp}
\usepackage{xcolor}
\usepackage{arydshln}
\usepackage{multirow}
\usepackage{multicol}

\usepackage{dsfont}

\newcommand{\ind}[1]{\mathds{1}\left\{#1\right\}}

\newcommand{\BR}{\mathbf{R}}
\newcommand{\BQ}{\mathbf{Q}}

\usepackage{mleftright}
\makeatletter
\newcommand*{\eval}{%
	\def\E@sub{}%
	\def\E@sup{}%
	\E@scripts
}
\newcommand*{\E@scripts}{%
	\@ifnextchar_\E@subscript{%
		\@ifnextchar^\E@supscript\E@finish
	}%
}
\def\E@subscript_#1{%
	\ifx\E@sub\@empty
	\def\E@sub{#1}%
	\else
	\errmessage{E already has a subscript}%
	\fi
	\E@scripts
}
\def\E@supscript^#1{%
	\ifx\E@sup\@empty
	\def\E@sup{#1}%
	\else
	\errmessage{E already has a superscript}%
	\fi
	\E@scripts
}
\newcommand*{\E@finish}[1]{%
	\mathbb{E}%
	\ifx\E@sub\@empty\else _{\E@sub}\fi
	\ifx\E@sup\@empty\else ^{\E@sup}\fi
	\mleft[#1\mright]%
}
\makeatother

\def\BibTeX{{\rm B\kern-.05em{\sc i\kern-.025em b}\kern-.08em
    T\kern-.1667em\lower.7ex\hbox{E}\kern-.125emX}}

\begin{document}

\title{Optimizing Risk-averse Human-AI Hybrid Teams\\
\thanks{This work was partially supported by the CHIST-ERA-19-XAI010 SAI project. M. Conti’s and A. Passarella's work was partly funded by the PNRR - M4C2 - Investimento 1.3, Partenariato Esteso PE00000013 - "FAIR" funded by the European Commission under the NextGeneration EU programme.}
}

\author{\IEEEauthorblockN{Andrew Fuchs}
\IEEEauthorblockA{\textit{Department of Computer Science} \\
\textit{Universit\`{a} di Pisa,} \\
\textit{National Research Council (CNR)}\\
Pisa, Italy \\
andrew.fuchs@phd.unipi.it}
\and
\IEEEauthorblockN{Andrea Passarella}
\IEEEauthorblockA{\textit{Institute for Informatics and Telematics}\\ % (IIT)} \\
\textit{National Research Council (CNR)}\\
Pisa, Italy \\
a.passarella@iit.cnr.it}
\and
\IEEEauthorblockN{Marco Conti}
\IEEEauthorblockA{\textit{Institute for Informatics and Telematics}\\ % (IIT)} \\
\textit{National Research Council (CNR)}\\
Pisa, Italy \\
marco.conti@iit.cnr.it}
}

\maketitle

\begin{abstract}
We anticipate increased instances of humans and AI systems working together in what we refer to as a hybrid team. The increase in collaboration is expected as AI systems gain proficiency and their adoption becomes more widespread. However, their behavior is not error-free, making hybrid teams a very suitable solution. As such, we consider methods for improving performance for these teams of humans and AI systems. For hybrid teams, we will refer to both the humans and AI systems as agents. To improve team performance over that seen for agents operating individually, we propose a manager which learns, through a standard Reinforcement Learning scheme, how to best delegate, over time, the responsibility of taking a decision to any of the agents. We further guide the manager's learning so they also minimize how many changes in delegation are made resulting from undesirable team behavior. We demonstrate the optimality of our manager's performance in several grid environments which include failure states which terminate an episode and should be avoided. We perform our experiments with teams of agents with varying degrees of acceptable risk, in the form of proximity to a failure state, and measure the manager's ability to make effective delegation decisions with respect to its own risk-based constraints, then compare these to the optimal decisions. Our results show our manager can successfully learn desirable delegations which result in team paths near/exactly optimal with respect to path length and number of delegations.
\end{abstract}


\section{Introduction}
\label{sec:intro}

The performance attainable by AI systems continues to improve and so we anticipate increased adoption to create scenarios where humans and AI systems, which we refer to both as agents, operate as a ``hybrid team.'' In a hybrid team, tasks can be offloaded to a single agent or there can be concurrent actions of multiple human and AI agents \cite{carroll2019utility,chen2018planning,10.1007/978-3-030-62056-1_44,westby2023collective,wu2021too}. As is evident from past examples \cite{agudo2024impact,cabrera2021discovering,haegler2010no,mahmood2022owning,reason1990human,russell2017human}, humans and artificial agents are certainly not error-free, so both humans and artificial agents are capable of erroneous decisions resulting from mistakes, risky behavior, etc. To ensure successful operation, considerations are needed to mitigate these issues. In some cases, this leads to the design of systems which abstain from decision-making when they have low confidence/experience \cite{afanador2019adversarial,chen2023learning}. Therefore, we utilize a manager, which oversees the operations of the hybrid team, and, over time, delegates the responsibility of taking a decision to either agent, based on the context of the specific problem, and the expected performance of agents at each specific point in time. The manager learns through observations of context and behavior how best to make delegation decisions. 

We highlight four primary features of our manager. First, clearly, the manager prioritizes selecting agents which provide higher chances of success, based on its own estimate of the performance of the individual agents at any point in time. Second, the measure of agent performance used by the manager is independent of those used to train the individual agents on the same task. This ensures the manager does not make unfair assumptions regarding access to private agent information or the manager being bound by the same measure of ideal behavior. Third, interventions of the manager (i.e., decisions on delegation choices) are also based on constraints regarding the behavior and outcomes of the team decisions. Typically, such constraints indicate how far (or close) the hybrid team is to severe conditions that may have catastrophic effects (e.g., an incident in case of driving scenarios). These constraints, together with estimates of the agents' actions, determine intervention points. Therefore, the manager can remain in the background until intervention is needed. Fourth, the manager learns how to optimize the performance of the team by choosing appropriate delegations and learns how to minimize the \emph{number} of such interventions. In realistic scenarios, managers that switch too frequently between team agents might not be practical, indeed.

We code these features into a standard Reinforcement Learning scheme, which we completely describe in \cite{fuchs2024optimizing}. Specifically, the manager reward model will receive reward-based feedback which includes penalties for delegations resulting in violation of constraints. The scale of the penalties will indicate how strongly averse the manager should be to making an intervention, which implicitly indicates aversion to those conditions which would lead to an intervention. This will serve to motivate the manager to recognize those agents which align best with the manager's constraints regarding behavior. Additionally, the manager reward will distinguish between task success and failure to ensure the manager also recognizes successful behavior regarding the underlying task. Combined, these aspects will incentivize a manager behavior policy which prioritizes successful team behavior and reduced frequencies of manager intervention.

While we have considered similar concepts in prior work \cite{fuchs2022cognitive,fuchs2023compensating,fuchs2023optimizing}, there are distinguishing factors for our model. First, in \cite{fuchs2022cognitive}, we propose a Reinforcement Learning (RL) model which mixes cognitively inspired Instance-Based Learning (IBL) \cite{aggarwal2014instance} and RL \cite{sutton2018reinforcement} to learn a delegation model based on human-inspired behavior understanding. At each time step, the manager selected the next delegated agent. We extended the delegation scenario in \cite{fuchs2023compensating} to cases of hybrid teams in driving contexts with impacted agent sensing. This method similarly relied on manager delegations at each step of an episode. Additionally, the manager observation was provided by combining the observations of the team agents. For our work in \cite{fuchs2023optimizing}, we focus instead on cases where agents may differ in the actions they can take in an environment. We required consistency for the states the agents can visit, but the diverse actions meant agents may move through the environment in diverse ways. For our new model \cite{fuchs2024optimizing}, as opposed to our prior works, our approach lets us isolate the manager model from the agent models. Unlike our previous cases, the manager has an entirely independent view of states and can only observe changes in state at intervention points. The manager will therefore ignore intermediary states, eliminating reliance on consistency between the manager and agent models. Further, we remove the previous expectation of a manager delegation at each time step to enable agent operation for windows of time steps. Additionally, we isolate the manager's notion of desirable behavior by eliminating assumptions of similarity between the manager and agent feedback regarding actions/behavior.

The main contribution presented in the paper is the analysis of the optimality for our delegation manager described in \cite{fuchs2024optimizing}, which supports the management of hybrid teams of diverse agents. Shifting from \cite{fuchs2024optimizing}, we introduce teams of agents which exhibit diverse behavior models according to their own notion of risk aversion. These agents are used to create teams for training and testing our manager model. We train our manager to identify agents which can successfully navigate the environment while avoiding violations of manager constraints. Unlike our scenario in \cite{fuchs2024optimizing}, we do not include cases of multiple agents moving concurrently in the environment; instead, we focus on demonstrating optimality for single teams with no additional agents, which prevents our environments from becoming non-stationary. The scenario, and limitation to a single team, enables comparison of our team to that of an optimal solution. From our tests, we see most cases resulting in a score at or near perfect for the manager.%, with worst case scores of at most four combined additional agent steps and interventions. The additional path lengths and/or interventions can be attributed to increased agent aversion to failure states, which results in longer paths than simply the shortest. Steps off an optimal path will typically result in at least one additional step to return to the shortest path, so each diversion will typically result in an increase of two to the score. The frequent scores near optimal indicate a strong performance from our manager and a clear indication of recognition with respect to ideal delegation decisions.

The remainder of this paper is as follows: Section~\ref{sec:related_work} outlines additional works of note which relate to and inform methods used in our framework; Section~\ref{sec:intervening_manager} introduces our manager model; Section~\ref{sec:grid_behavior} outlines the demonstration scenario and related details; Section~\ref{sec:results} defines our test scenarios and provides results; and Section~\ref{sec:conclusion} concludes the work.

\subsection{Related Work}\label{sec:related_work}
\subsubsection{Hybrid Reinforcement Learning}
Hierarchical Reinforcement Learning (HRL) represents a selection of methods and features which inspired our approach. HRL approaches will typically elicit models represented in a hierarchical manner. The lower-level policies operate at more granular levels of actions and state transitions. At these lower levels, models of behavior dictate outcomes for sequences of decisions and higher-level policies can learn when to select these policies. The behaviors at the lower level are referred to as ``options,'' which execute until a termination is triggered. In \cite{garcia2020learning}, the proposed approach demonstrates a method to learn ``reusable'' options, where new options are added if it is apparent, they may provide improved performance. Regarding the shift from one option to the next, \cite{erskine2022developing} demonstrates an approach for considering the compatibility of successively selected options, which can help to reduce the complexity of making the transition from an option to its successor. While relevant, our model assumes previously learned behavior policies for the team agents and no hierarchical layers of abstraction, so we are not following the common HRL approach of learning policies at various levels.

As an extension, HRL concepts can also be applied in scenarios of multiple concurrently operating agents \cite{chakravorty2019option,chen2022multi,kurzer2018decentralized,lau2012coordination,menda2018deep,rohanimanesh2002learning,singh2020hierarchical,yang2022ldsa}. In the multi-agent scenario, agents can operate simultaneously while attempting to accomplish potentially distinct tasks. As seen in \cite{yang2022ldsa}, cooperative or compatible tasks can allow knowledge sharing to help solve compatible tasks. The use of multiple agents also affords a measure of compatibility, so both agent and task can be considered to ensure the best alignment of agents to tasks. With our assumption of multiple agents but a single acting agent, we will therefore not be making direct comparison to the multi-agent approaches. For further distinction between our model and HRL approaches, we refer to Section~\ref{sec:intervening_manager}.

\subsubsection{Delegation}
In addition to HRL methods, we also note the relation of our approach to delegation \cite{fuchs2022cognitive,fuchs2023compensating,fuchs2023optimizing,jacq2022lazy,balazadeh2020switch,straitouri2021triage,jacq2022lazy,richards2016delegate,palmer2020assessing,candrian2022rise}, where a manager is similarly tasked with identifying adequate agents for a task given the current state and some measure of agent desirability (e.g., performance, cost, etc.). Commonly, the manager is tasked with making decisions based on a measure of agent performance and their corresponding cost of executing actions. In this model, the manager will require a more detailed and deeper understanding of desirable agent behavior and measure of agent cost. As an example, \cite{straitouri2021triage} demonstrates a delegation model using performance or priority values to train a Reinforcement Learning model to operate under algorithmic triage. Similarly, \cite{balazadeh2020switch} demonstrates the use of costs associated with scenario features such as agent operation or switching costs in the manager reward. This idea can also be linked to \cite{jacq2022lazy} as it demonstrates a case where the delegation policy maintains a bias toward a lower-cost policy when possible and switches to a higher-cost and more performant policy when necessary. As we focus on a manager model with fewer assumptions regarding access to these features, our approach includes distinct aspects from the highlighted delegation models. Again, our manager will instead be attempting to reduce the need to make such delegation decisions while also not requiring access to an explicit measure of action values for the team agents.

\section{Intervening Manager}\label{sec:intervening_manager}
We model our manager through a modified Markov Decision Process (MDP) and a series of Absorbing Markov Chains (AMC) (see \cite{fuchs2024optimizing} for more details on their general features). Specifically, let $s$ be a state in the state space $S$ of the MPD describing the manager's behavior. We define a function $\beta(s): S\rightarrow\{0,1\}$ to identify intervention states (i.e., states where the manager must evaluate a possible change of delegation) as those states where $\beta(s)=1$ holds true. We model the operations of the agent delegated between two consecutive intervention points as a Markov Chain, where we identify a subset of states $S_\BQ$ that do not generate a manager intervention (i.e., for which $\beta=0$), and a set of states $S_\BR$ where interventions are made.

Between two consecutive delegations, we model this as an Absorbing Markov Chain (see \cite{fuchs2024optimizing}) where $S_\BQ$ and $S_\BR$ are transient and absorbing states, respectively. When the manager decides on the next delegation, a ``new" Markov Process starts, whose absorbing states are those that will trigger the next delegation action by the manager. As such, the manager MDP observes only the absorbing states of the underlying AMCs, as those are the ones where the manager takes decisions, and thus, as explained in the following, rewards are provided based on a standard Reinforcement Learning approach. Finally, in the following, we define an episode as the sequence of actions from the beginning of the task to its completion, which may even result in success of the hybrid team (``goal found") or not (``goal not found").

More formally, to model the dynamics for our manager, we use a modified MDP which refer to as an Intervening MDP (IMDP) $M_I = \langle S=S_\BR\bigcup S_\BQ, A_M, R_M, T_M, D, \beta, \gamma\rangle$ with
\begin{itemize}
    \item $S$, the MDP state space
    \item $S_\BR\subseteq S$, the set of intervention states
    \item $S_\BQ\subseteq S$, the set of unobserved delegated operation states
    \item $A_M = \{i\in D\}$, the manager agent selection action space
    \item $R_M$, the manager's reward function
    \item $T_M$, the manager's transition function
    \item $D$, the set of agents available for delegation
    \item $\beta$, the intervention cue function
\end{itemize}

Moreover, the reward function of the manager is defined as follows (assuming it is computed at the end of an episode):
\begin{equation}\label{eqn:manager_reward}
	R_M=\begin{cases}
	    1 - \tanh\left(\nu\cdot\rho\right),&\quad\textrm{goal found}\\
            -\tanh\left(\nu\cdot\rho\right),&\quad\textrm{otherwise}
	\end{cases}
\end{equation}
where $\rho$ is the ``cue frequency," i.e., the number of times interventions have been triggered during the observed episode, and $\nu$ is a scaling coefficient, which we use to scale how strongly the penalties degrade the episode's value. Note that, using $\tanh()$ allows us to maintain a bound on the scale of the penalty for RL training.

The manager IMDP will make transitions from $s_i\in S_\BQ$ to $s_j\in S_\BR$ based on the policy of the delegated agent $\pi_d$ for $d\in D$ and $\beta$. Therefore, with an infinite horizon, the manager transitions can be modeled as
\begin{equation}
    P(s_j|s_i,d) = b^{\pi_d}_{ij}
\end{equation}
with $b^{\pi_d}_{ij}$ referring to the transition probabilities of the underlying AMC between two interventions, under the specific policy $\pi_d$ of the agent delegated during that time window. The use of a series of AMCs implies that we must allow an intermediary transition from an intervention state to the start of a new delegation window. This will implicitly occur in the same state, so we extend the definition of $\beta$ to accommodate this aspect. The extension allows the representation to shift from an absorbing state that does not terminate an episode to a recurrent state which starts a new AMC chain. In this case, the model becomes
\begin{equation}
    \beta(s, \eta_s) = \begin{cases}
        1, \quad (s \textrm{ violates constraints} \wedge \eta_s = 0 ) \vee s=s_g\\
        0, \quad s \textrm{ satisfies constraints} \vee \eta_s = 1
    \end{cases}
\end{equation}
where $\eta_s$ is an indicator denoting whether a new delegation was made in state $s$. With this $\beta$, a delegation converts the current state into the starting state of a new AMC which continues until they either reach a goal/terminal state or a new absorbing state.

The final aspect of the manager model, according to a standard RL scheme, is the definition of its value function. Specifically, the manager delegation policy $\pi_m$ gives a value function for state $s_j \in S_Q$
\begin{equation}\label{eq:manager_V}
    V^{\pi_m}(s_j) = \sum_{d\in D}\pi_m(d|s_j)\sum_{s_k\in S_\BR} b^{\pi_d}_{jk}\left[R_M + \gamma V^{\pi_m}(s_k)\right]
\end{equation}
where $b^{\pi_d}_{jk}$ allows the manager's value function to depend on the AMC transitions according to the delegation to agent $d$. The value estimate is based on manager delegation policy $\pi_m$ and states observed according to delegated agent behavior policies $\pi_d$. The model can be defined recursively to estimate expected state value with respect to the probability of delegation decision $\pi_m(d|s)$, probability of next observed state $b^{\pi_d}_{jk}$ following delegated agent policy  $\pi_d$, and the discounted next state value $\gamma V^{\pi_m}(s_k)$. The discount $\gamma$ indicates the weight for the estimated value of future states. The manager estimates the value of states according to its behavior policy via \eqref{eq:manager_V}, which estimates the expected rewards for trajectories leading from the current state $s_j$. These rewards, immediate or episodic, are dictated by the manager reward function $R_M$ and indicate the value/reward for each decision made by the manager. For immediate rewards attributed at each intervention, the manager will observe in \eqref{eq:manager_V} the immediate feedback resulting from the transition to state $s_k$. For the episodic case (i.e., when rewards are attributed only at the end of an episode), the manager observes a delayed reward based on all delegations occurred during the episode. This is done by backtracking the observed rewards and assigning these values to the explored states. Both cases of feedback are consistent with standard RL methods. 

We can therefore derive the optimal value function $V^*$ as
\begin{equation}
    V^*(s_j) = \max_d\sum_{s_k\in S_\BR}b^{\pi_d}_{jk}\left[R_M + \gamma V^*(s_k)\right]
\label{eq:optimal_V}
\end{equation}
and state-action value function $Q^*$
\begin{equation}
    Q^*(s_j,d) = \sum_{s_k\in S_\BR}b^{\pi_d}_{jk}\left[R_M + \gamma V^*(s_k)\right]
\label{eq:optimal_Q}
\end{equation}

As shown in \cite{fuchs2024optimizing}, our model can be shown to conform to standard RL requirement for convergence. Further, we can utilize RL techniques to learn an optimal manager policy $\pi_m$.

\section{Specific scenario for optimality analysis}\label{sec:grid_behavior}
\begin{figure}[htbp]
    \begin{center}
        \begin{subfigure}[t]{0.17\textwidth}
            \includegraphics[width=\textwidth]{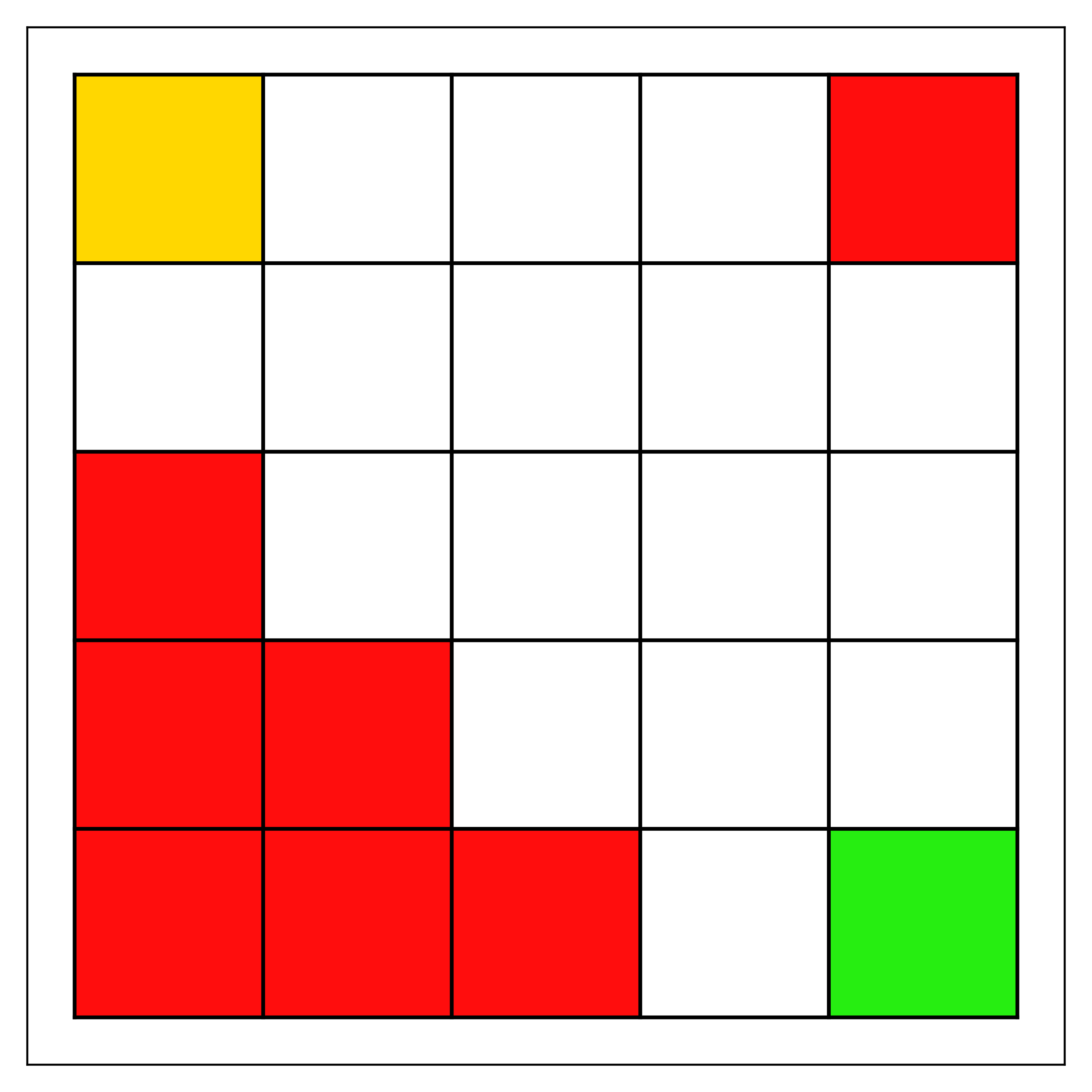}
            \caption{Angle Cliff}
            \label{fig:angle_cliff}
        \end{subfigure}
            \hspace{3mm}
        \begin{subfigure}[t]{0.17\textwidth}
            \includegraphics[width=\textwidth]{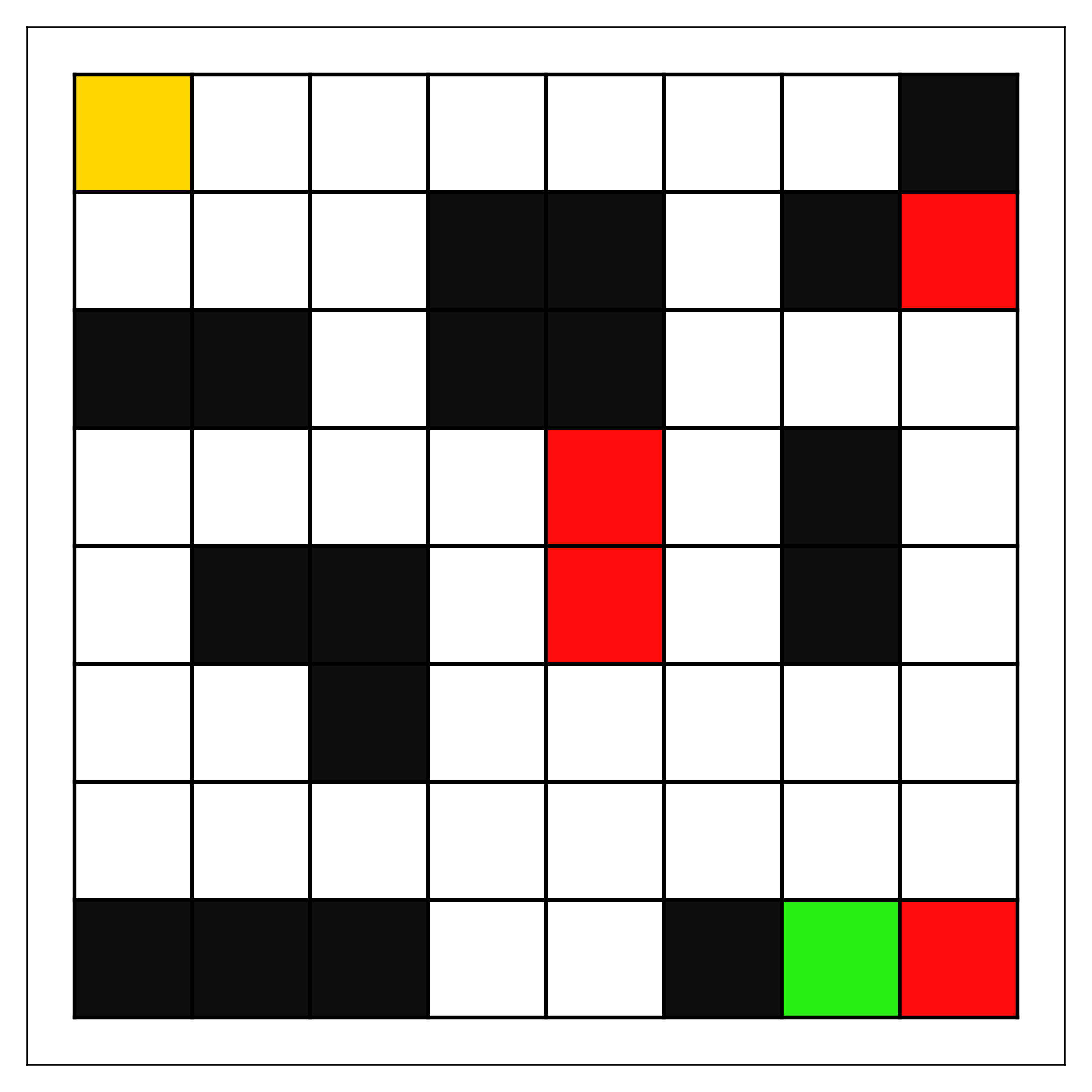}
            \caption{Maze}
            \label{fig:maze}
        \end{subfigure}\\
        \begin{subfigure}[t]{0.4\textwidth}
            \includegraphics[width=\textwidth]{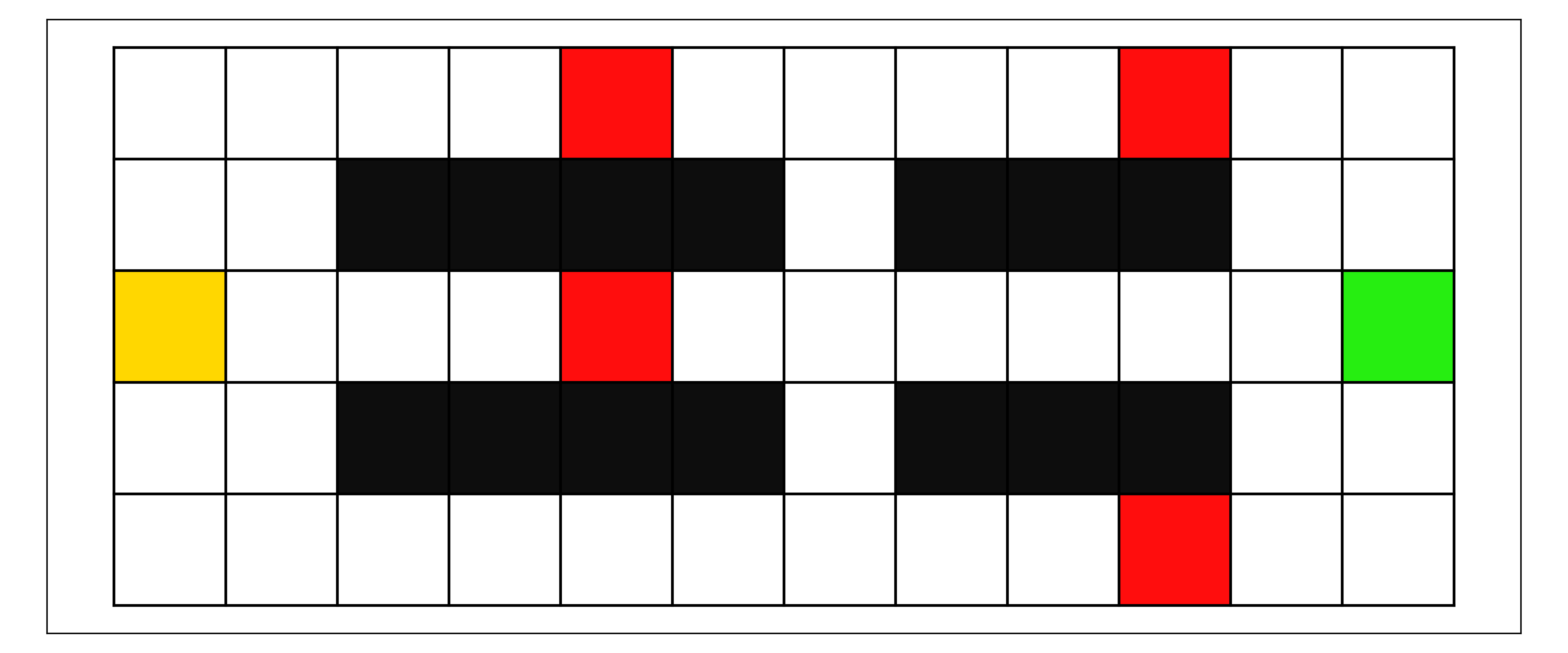}
            \caption{Hallways}
            \label{fig:hallways}
        \end{subfigure}
        \caption{Grid environments. States: Yellow - Start, Green - Goal, White - Open, Black - Wall, and Red - Failure.}
        \label{fig:grid_environments}
    \end{center}
\end{figure}

We consider a specific scenario, described hereafter, which is particularly suited to analyze the performance of a manager defined as explained in Section~\ref{sec:intervening_manager} with respect to the ideal optimal behavior. Specifically, we utilize teams of agents which are tasked with navigating a grid environment (see Fig.~\ref{fig:grid_environments}) by moving from a start state to a goal state. Grid environments are defined by a collection of grid cells which the agents navigate by selecting actions denoting the direction they wish to travel (i.e., left/right/up/down). The actions will move the agents to an adjacent cell in the direction of the action selected. If the agent chooses to move in the direction of a wall or boundary, by the nature of the grid environment, the boundary prevents movement in that direction and the agent will remain in the same state. We augment the grid environment by adding a state type which we denote as a ``failure'' type state. If an agent enters either a goal or failure state, then the episode will terminate.

The failure states provide a means to guide manager behavior and define manager constraints. For the management of agents navigating a grid environment, our manager will have constraints defining the lowest accepted proximity to a failure state. If the team violates the proximity constraint, then the manager's $\beta$ function will indicate that an intervention is required, and a delegation decision will be made. With the characteristics of the grid environment, we will determine distance according to the Manhattan Distance
\begin{equation}\label{eq:manhattan_dist}
    \delta(s_1, s_2) = |x_1 - x_2| + |y_1 - y_2|
\end{equation}
for cells $s_i = (x_i, y_i)$ and $(x_i, y_i)$ indicating the column and row of the cell, respectively. We use the Manhattan distance as this corresponds directly to the number of steps needed to reach a failure state given there are no boundaries. As such, we get an estimate of the importance of a nearby failure state.

The use of distance-based notions of optimality and safety will allow us to directly observe the impact manager oversight has on the team's behavior. We would expect to see cases where the manager learns delegation selections which balance the team's risk, with respect to manager constraints, and the performance measure the agents used to learn. The manager will of course not observe these performance measures, but will observe how the learned behaviors of the agents have been impacted while learning according to these performance measures. Combined, this allows us to measure these factors to indicate how well the team and manager can work together to accomplish their task. We will discuss these aspects, and how we measure them, in Section~\ref{sec:manager_task}.

\subsection{Navigating Agent Training}\label{sec:navigating_agent_task}
To build teams of agents, we train the navigation agents independently to generate behavior policies for each agent. Once trained, each behavior policy defines a candidate agent for a team. For the training, we represent the grid environment as an MDP. The transitions operate as described above to support movement between adjacent states and agents receive rewards to indicate the value of their action choices. In our independently trained agent models, we provide rewards to motivate desirable behavior. For the navigation task, we split the reward into to two components to give observed reward
\begin{equation}
    r = R(s,a,s') + R\left(\min_{s_f}\delta(s, s_f)\right)
\end{equation}
where $s$ is the current state and $s'$ is the next state reached resulting from the action $a$ in state $s$. First, the agents receive a penalty each time they are within an agent-specific distance of a failure state $s_f$. The scale of the penalty depends on agent distance and is set for each agent, given by
$R(\delta)\in\mathbb{R}$ according to
\begin{equation}
    R\left(\min_{s_f}\delta(s, s_f)\right) = r_d
\end{equation}
for $\delta$ from Equation~\ref{eq:manhattan_dist}. As we are generating unique agents, each agent $d$ will have its own set of values for $r_d$ given as a parameter for agent training. For example, with closest failure state $s_f$, an agent could observe
\begin{equation}\label{eq:dist_penalty}
    R(\delta(s, s_f))=\begin{cases}
        -20,&\min_{s_f}\delta(s, s_f)=1\\
        -10,&\min_{s_f}\delta(s, s_f)=2\\
        \quad~0,&\textrm{otherwise}
    \end{cases}
\end{equation}
to give unique penalties for $\delta(s, s_f)<3$ and no penalty for $\delta(s, s_f)\geq 3$. Also, once the agent has moved to a cell such that $\min_{s_f}\delta(s, s_f)\geq 3$, there will be no observed proximity penalty.

In addition to the distance-based penalty, we provide rewards aligning to the general navigation task. These rewards motivate shortest paths and eliminate erroneous behavior(s) (e.g., collision with a wall/boundary). These rewards are given according to
\begin{equation}\label{eq:navigation_reward}
    R(s,a,s')=\begin{cases}
        -10,&\textrm{wall/boundary collision}\\
        -20,&\textrm{$s'$ is a failure state}\\
        100,&\textrm{$s'$ is a goal state}\\
        -1,&\textrm{otherwise}
    \end{cases}
\end{equation}
to motivate agents to learn efficient paths to the goal.

\begin{figure}[htbp]
    \begin{center}
        \begin{subfigure}[t]{0.17\textwidth}
            \includegraphics[width=\textwidth]{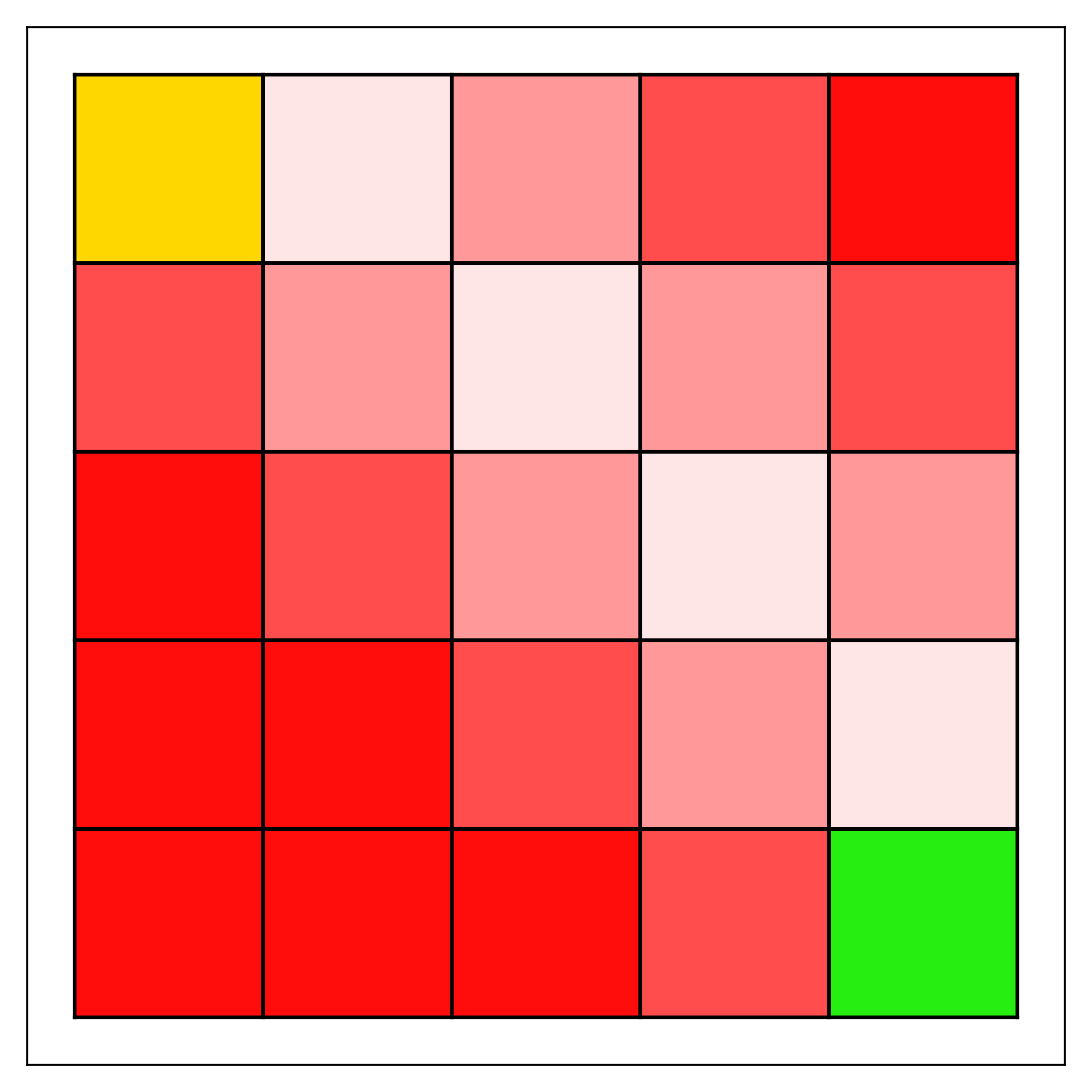}
            \caption{Angle Cliff}
            \label{fig:angle_cliff_3}
        \end{subfigure}
        \hspace{3mm}
        \begin{subfigure}[t]{0.17\textwidth}
            \includegraphics[width=\textwidth]{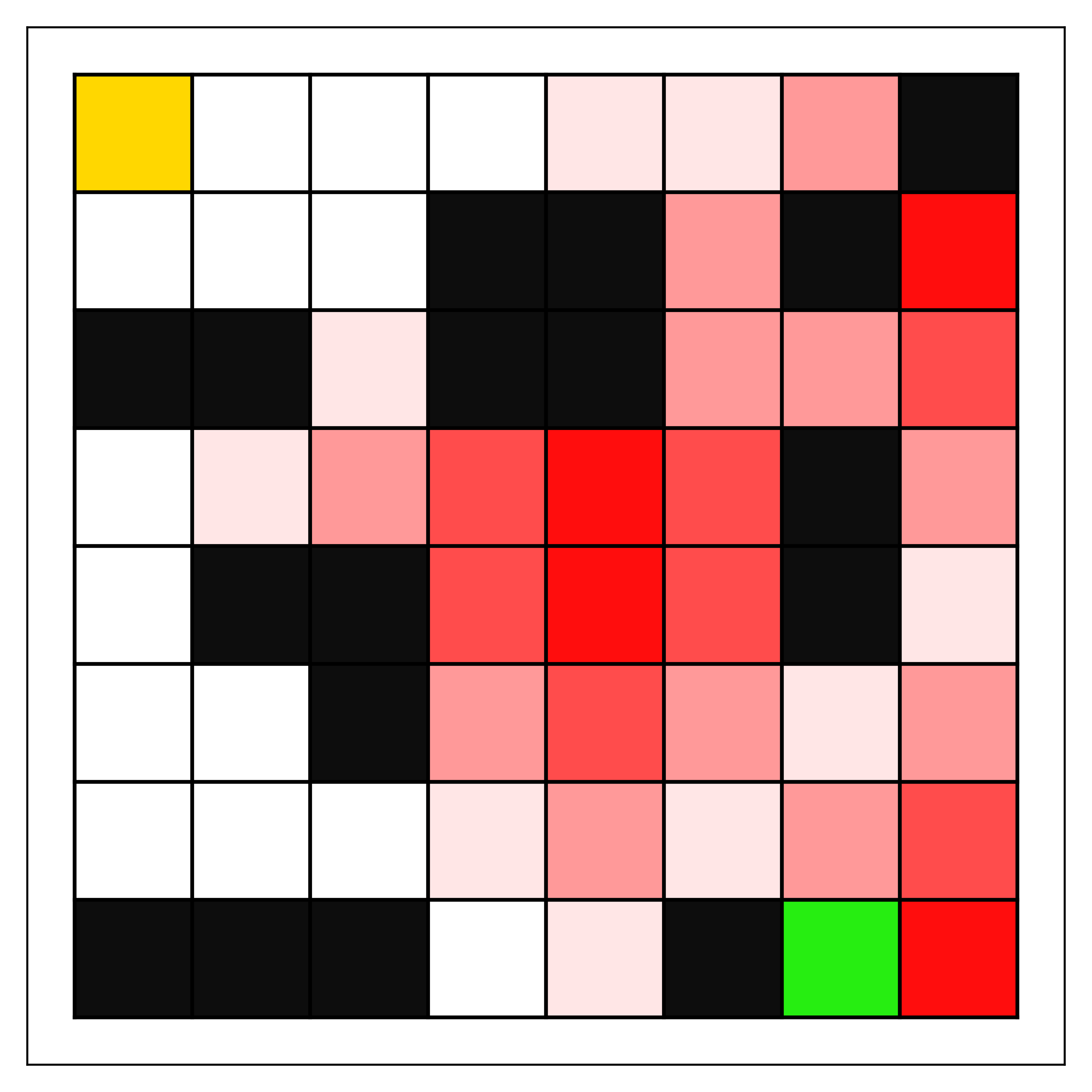}
            \caption{Maze}
            \label{fig:maze_3}
        \end{subfigure}\\
        \begin{subfigure}[t]{0.4\textwidth}
            \includegraphics[width=\textwidth]{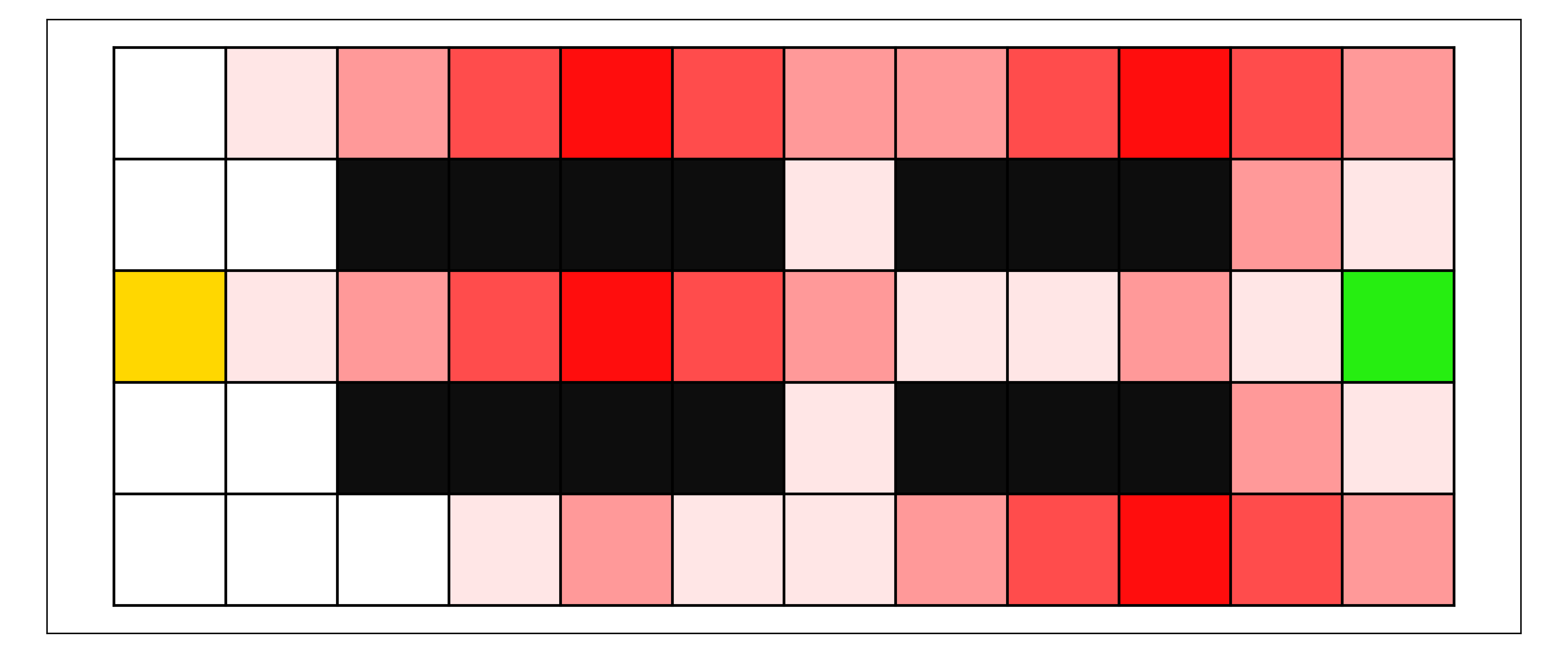}
            \caption{Hallways}
            \label{fig:hallways_3}
        \end{subfigure}
        \caption{Grid cell aversion level when safe distance is at Manhattan Distance of three. Red cells indicate the failure states and lighter shades of red/pink indicate the severity by which agents should avoid the corresponding cell in a path.}
        \label{fig:grid_environments_with_dists}
    \end{center}
\end{figure}

The distance-based penalty will therefore determine how close an agent is likely to reach a failure state while learning an efficient path. As an example, we refer to Fig.~\ref{fig:grid_environments_with_dists}, where we demonstrate the significance placed on cells near failure states when the safe distance is three. In the figure, there is a clear pattern of degrading importance placed on cells as the distance to a failure state increases. Therefore, based on the severity of these penalties, an agent's path will prioritize the cells which are of least concern with respect to the safety distance (e.g., the cells furthest left in Fig.~\ref{fig:maze_3}). It is important to note that this does not assume any compatibility or similarity to the manager's constraints. This allows us to create ensembles of agents where each act according to their previously learned model and corresponding level of accepted risk.

\subsection{Manager Task}\label{sec:manager_task}

As defined for our framework, the agents under manager oversight will operate by choosing actions until there is need for an intervention. In the grid navigation case, from the manager's perspective, this will amount to teams making errors by moving to a state which is too close to a failure state (or entering a failure state), according to the manager's constraints defining $\beta$ regarding the safe distance.

Similar to Equation~\ref{eq:dist_penalty}, the manager will use a $\beta$ function based on $\min_{s_f}\delta(s, s_f)$ for current state $s$. In our scenario, the distance is used to define $\beta$ such that
\begin{equation}
    \beta(s,\eta_s) = 1 \Leftrightarrow \min_{s_f}\delta(s, s_f) \leq \delta_I \wedge \eta_s = 0
\end{equation}
where $\delta_I$ sets the manager's limit on proximity. Thus, we see the manager will not intervene until the minimum Manhattan distance meets the manager's threshold. This ensures the manager bases its intervention decision on a separate and compatible measure of safety.

Following Equation~\ref{eqn:manager_reward}, the manager will observe a reward which indicates task success or failure, with the scale of the reward being dictated by the number $\rho$ of interventions triggered. As defined, the reward values are not directly dependent on the safety distance $\delta_I$; instead, the manager reward will be based on intervention frequency. Therefore, there is an implicit dependence on $\delta_I$, but the key factor is the behavior of the delegated agents and the team's success regarding the task.

As we isolate the manager and agent views on safety, the team dynamics will vary accordingly. Agents selected by the manager will operate according to their own acceptable safety distance, which will guide their actions and resulting paths. Similarly, the manager’s constraints on safe behavior will determine how suitable an agent is for delegation. Teams composed entirely of agents which accept more risk than the manager will typically result in frequent manager intervention while agents with more restrictive risk aversion will lead to few/no needed interventions by the manager. Consequently, the manager's performance will depend on agents available, how the agents choose to operate, and how well the manager can associate these factors to learn desirable behavior.

Given our view of the manager's impact on team performance, the notion of optimal manager behavior in our scenario will be that of the sum of path length and number of manager interventions. The focus on path length will indicate the kind of path the team finds according to their risk aversion levels, while the focus on interventions clearly indicates how well the team aligns with the manager's constraints. With that in mind, the performance for our manager will be measured as
\begin{equation}\label{eqn:manager_cost}
    c = m + n_\beta
\end{equation}
for path length $m$ and
\begin{equation}
    n_\beta = \sum_{s_t\in\{s_1,\dots,s_m\}} \ind{\beta(s_t, \eta_{s_t}) = 1}
\end{equation}
denoting the number of interventions for an episode trajectory $\{s_1,\dots,s_m\}$. The $c$ values for each test episode are then averaged to calculate the mean performance $\bar{c}$ for a managed team. We perform the tests by running 50 episodes per two agent team in each environment and intervention distance case.

To generate bounds on performance, we measure the optimal cost in an environment with consideration for how the values of $m$ and $n_\beta$ are related. These values will be determined by the specific grid environment and manager constraints. We use a shortest path algorithm to calculate the minimal value possible for $c$ for a given manager. The measure of a shortest path is augmented to accommodate the manager's constraints. Any path which would induce an intervention will have its weight increased by one each time $\beta=1$ to indicate an intervention penalty. Hence, the shortest path algorithm will return a path which has a minimal sum of grid cells traversed and manager interventions incurred.

In cases where the manager's performance score deviates from optimal, this can result from three outcomes. First, the team may have only agents with aversion higher than the manager, so their paths are longer than the shortest possible. For example, consider an agent avoiding the central failure state in Figure~\ref{fig:maze_3}. A path avoiding this state will incur additional steps which move away from the goal and toward the left boundary before turning back toward the goal. A shorter path is possible adjacent to the failure state, so a team that avoids this path could have a sub-optimal score. Second, the manager could choose an agent which has not gained much experience in a part of the grid due to its training not performing much exploration in that region. If the manager has a team of agents which enters such a region, then there could be slight confusion and resulting longer paths until the team returns to the desirable path. Last, the manager could simply choose the sub-optimal agent in a state and result in less desirable paths than if the manager chose the other agent.

\begin{figure}[t]
    \centering
    \includegraphics[width=0.9\linewidth]{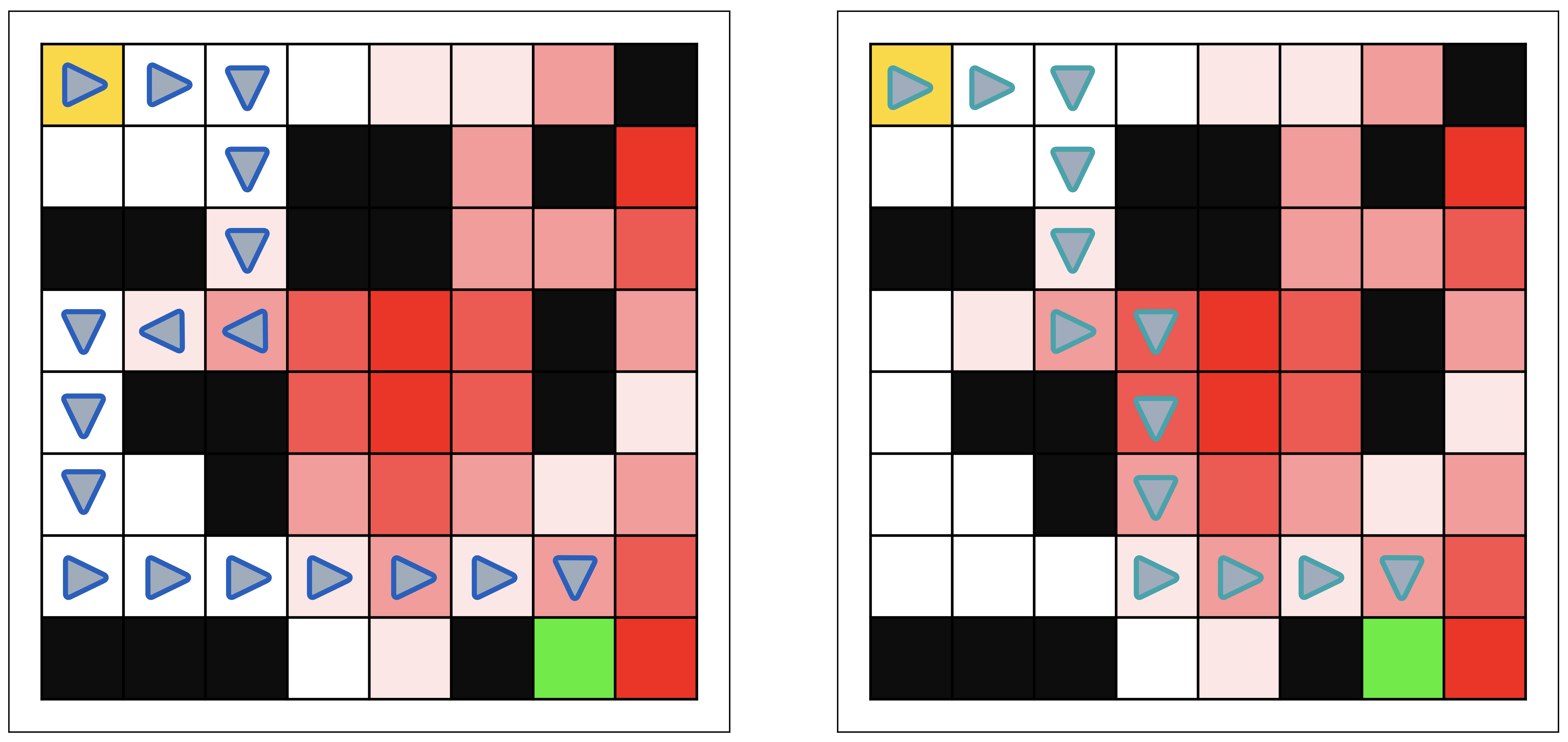}
    \caption{Comparison of agent paths according to risk aversion level. Left image indicates strong aversion while right image indicates a higher acceptable risk.}
    \label{fig:maze_paths}
\end{figure}

For concrete examples, given our measure of manager performance, we will compare the $c$ values to a measure of optimal paths. %In this case, we will view optimality according to the smallest sum of path length and interventions. This measure will indicate the best possible value a team could achieve given a comparable notion of safety. In other words, teams with at least one agent with equivalent risk aversion to the manager should select a path which achieves this minimum. On the other hand, teams with agents with lower risk aversion would likely result in more manager intervention as the agents will be more willing to reach states closer to the failure states.
As indicated in Figure~\ref{fig:maze_paths}, agent risk aversion levels will determine the path an agent uses to traverse a grid environment. On the left side, we see an agent with high penalties for $R(\delta(s, s_f))$ (e.g., $R(1) = -35$ and $R(2) = -15$), while the right side shows behavior for lower values (e.g., $R(1) = -15$ and $R(2) = -5$). As indicated, the strength of the aversion penalties can have a significant impact on the path length for an agent/team. When paired with a manager, the agent and manager aversion levels will impact the overall resulting paths. Agents with high penalties, such as the left side of Figure~\ref{fig:maze_paths}, will have fewer cases where their path is close enough to a failure state to trigger a manager intervention. Therefore, the best $c$ a manager with $\delta_I\leq 2$ can achieve given a team of these agents will be bounded by their path length (i.e., $c=18$). On the other hand, a manager with a team of agents selecting paths such as the right side of Figure~\ref{fig:maze_paths}, will have higher dependence on $\delta_I$. With $\delta_I=1$, the team will trigger an intervention and so $n_\beta=1$ and $m=14$, giving a best possible score $c=15$. Hence, manager score depends on both their ability to select suitable agents and the best possible behavior any agent could achieve. Therefore, our score of optimality indicates how well the manager can achieve these values and how the team impacts the manager's score.

\section{Results}\label{sec:results}

We train and test our managers across several grid environments and risk aversion levels. The grid environments (see Fig.~\ref{fig:grid_environments}) are designed to include varying levels of navigation difficulty and ease of failure state avoidance. Further, we generate several navigation agents with varying levels of risk aversion (see Section~\ref{sec:navigating_agent_task}) to create diverse teams for the manager. Given the agent parameters and training, all team agents are trained to avoid entering the failure state, but they are distinct in their risk feedback (i.e., $R(\delta)$).

As noted, the performance of the manager is compared to an estimate of the optimal path possible, according to the grid and corresponding manager distance-based error threshold. The navigating agents, by definition, are motivated to find paths which minimize path length and failure state proximity penalties, so the team's path will serve as an indicator of manager performance. Given that each agent is trying to reach the goal in the shortest path that does not violate their own view of risk, the agent a manager chooses will determine the length of path the team finds. Consequently, the manager's delegation decisions will directly impact how short a path is taken by the team. If the manager selects an agent with a higher aversion level than itself, the team may take more steps to reach the goal than if an agent with lower aversion were selected. Hence, manager performance can be viewed according to the manager's impact on team paths and intervention frequency.

In Tables~\ref{tab:angle_cliff}-\ref{tab:hallways}, we demonstrate the manager performance in our various grid environments. The manager training was performed across teams with a mixture of risk-aversion levels and manager intervention distances $\delta_I\in[0, 1, 2, 3]$. The inclusion of a zero-distance intervention serves as a base level to indicate how a team can perform when there is no intervention without reaching a terminal error state. Note, the navigating agents we trained have a success rate of 100\% with respect to their reaching the goal state at test time. This means the manager only needs to consider agent performance and $\beta$ with respect to the cost function defined in Equation~\ref{eqn:manager_cost}. Agents avoiding failure states ensures our comparison of results is consistent across the various teams and there is no ambiguity caused by the appearance of shorter paths due to failure paths.
% 
% In cases where the manager score deviates from optimal, this can result from two outcomes. First, the team may have only agents with aversion higher than the manager, so their paths are longer than the shortest possible. For example, consider an agent avoiding the central failure state in Figure~\ref{fig:maze_3}. A path avoiding this state will incur additional steps which move away from the goal and toward the left boundary before turning back toward the goal. A shorter path is possible adjacent to the failure state, so a team that avoids this path could have a sub-optimal score. Second, the manager could choose an agent which has not gained much experience in a part of the grid due to its training not performing much exploration in that region. If the manager has a team of agents which enters such a region, then there could be slight confusion and resulting longer paths until the team returns to the desirable path.

\begin{table}[t]
    \begin{center}
        \caption{Intervention results for managed team in Angle Cliff.}\label{tab:angle_cliff}
        \begin{tabular}{|c|c|c|c|c|}
            \hline
            \multirow{2}{*}{\textbf{Team Types}} & \multicolumn{4}{c|}{\textbf{Intervention Distance}}\\\cline{2-5}
             & 0 & 1 & 2 & 3 \\
            \hline
             None, Low & 8.0 (8.0) & 9.55 (8.0) & 13.14 (12.0) & 15.0 (15.0)\\
            \hline
             None, Medium & 8.0 (8.0) & 9.50 (8.0) & 13.31 (12.0) & 15.0 (15.0)\\
            \hline
             None, High & 8.0 (8.0) & 9.54 (8.0) & 13.39 (12.0) & 15.0 (15.0)\\
            \hline
             Low, Medium & 8.0 (8.0) & 8.0 (8.0) & 12.0 (12.0) & 15.0 (15.0)\\
            \hline
             Low, High & 8.0 (8.0) & 8.0 (8.0) & 12.0 (12.0) & 15.0 (15.0)\\
            \hline
             Medium, High & 8.0 (8.0) & 8.0 (8.0) & 12.0 (12.0) & 15.0 (15.0)\\
            \hline
        \end{tabular}\\
        \bigskip
        \caption{Intervention results for managed team in Maze.}\label{tab:basic_8x8}
        \begin{tabular}{|c|c|c|c|c|}
            \hline
            \multirow{2}{*}{\textbf{Team Types}} & \multicolumn{4}{c|}{\textbf{Intervention Distance}}\\\cline{2-5}
             & 0 & 1 & 2 & 3 \\
            \hline
             None, Low & 14.9 (14.0) & 15.5 (15.0) & 18.5 (18.0) & 21.0 (21.0)\\
            \hline
             None, Medium & 15.2 (14.0) & 15.5 (15.0) & 18.5 (18.0) & 21.0 (21.0)\\
            \hline
             None, High & 15.0 (14.0) & 15.4 (15.0) & 18.9 (18.0) & 21.8 (21.0)\\
            \hline
             Low, Medium & 17.0 (14.0) & 17.0 (15.0) & 19.0 (18.0) & 23.0 (21.0)\\
            \hline
             Low, High & 17.0 (14.0) & 17.0 (15.0) & 19.0 (18.0) & 23.0 (21.0)\\
            \hline
             Medium, High & 17.0 (14.0) & 17.0 (15.0) & 19.0 (18.0) & 23.0 (21.0)\\
            \hline
        \end{tabular}\\
        \bigskip
        \caption{Intervention results for managed team in Hallways.}\label{tab:hallways}
        \begin{tabular}{|c|c|c|c|c|}
            \hline
            \multirow{2}{*}{\textbf{Team Types}} & \multicolumn{4}{c|}{\textbf{Intervention Distance}}\\\cline{2-5}
             & 0 & 1 & 2 & 3 \\
             \hline
             None, Low & 15.0 (15.0) & 15.0 (15.0) & 18.0 (18.0) & 24.5 (24.0)\\
            \hline
             None, Medium & 15.0 (15.0) & 15.0 (15.0) & 18.0 (18.0) & 24.5 (24.0)\\
            \hline
             None, High & 15.0 (15.0) & 15.0 (15.0) & 18.0 (18.0) & 24.5 (24.0)\\
            \hline
             Low, Medium & 15.0 (15.0) & 15.0 (15.0) & 18.0 (18.0) & 24.5 (24.0)\\
            \hline
             Low, High & 15.0 (15.0) & 15.0 (15.0) & 18.0 (18.0) & 24.5 (24.0)\\
            \hline
             Medium, High & 15.0 (15.0) & 15.0 (15.0) & 18.0 (18.0) & 24.5 (24.0)\\
            \hline
        \end{tabular}
    \end{center}
\end{table}

We present the results in the tables in the following format. First, the ``Team Types'' indicates the level of aversion for a team agent with respect to proximity to a failure state: None, Low, Medium, and High. These aversion levels correspond to the maximum $\delta(s,s_f)$ such that $R(\delta(s, s_f))<0$, with $\textrm{None}\rightarrow0$, $\textrm{Low}\rightarrow1$, $\textrm{Medium}\rightarrow2$, $\textrm{High}\rightarrow3$. Remember that agent paths will vary according to these aversion levels (see Fig.~\ref{fig:grid_environments_with_dists}). In the remaining columns, we demonstrate the team's performance under managers with varying intervention distances. The team/manager scores are given as the team score followed by the estimated optimal score in parenthesis. The score is calculated according to combined shortest path and intervention cost $c$ defined in Section~\ref{sec:manager_task}. 

In Table~\ref{tab:angle_cliff}, we see the results for the Angle Cliff environment. In this grid, the team will have few chances to avoid intervention cases when $\delta_I$ increases. Still, there is a path along the diagonal that agents can traverse to minimize these interventions (see Fig.~\ref{fig:angle_cliff_3}). As indicated by the results, the manager can identify the sequence of agents which generates results nearing these optimal paths. As the agents were trained to find a path that fits their view of acceptable risk, without a prescribed training from any starting state, we can attribute some of the sub-optimal paths (e.g., ``None, High'' team with Intervention Distance: 2) to either risk aversion or minor agent confusion when the team enters less-frequently experienced states. In other words, some delegation changes and sub-optimal team agent behavior could result in a path which diverges slightly from a policy-agnostic shortest path. This could lead to backtracking (e.g., ``right'' action followed by a ``left'' action) to return to the optimal path, which increases $c$ by one or more. Further, any single deviation from the diagonal will result in the next states incurring additional penalties, so small deviations from the center path will result in these minor deviations in team score. On the other hand, for teams where the agents are of similar risk-aversion level to the manager (e.g., ``Low, Medium'' and $\delta_I=1$), we see indications of good alignment between the manager and team. In these cases, the scores align well with our measure of optimal as the team paths follow trajectories which align with the manager's constraints and lead to optimal results.

For the next results, Table~\ref{tab:basic_8x8} indicates the team performance in the Maze grid. As noted in Section~\ref{sec:manager_task}, this environment offers an example of cases where the agent aversion to risk could have significant impact on the paths taken. %More specifically, agents with higher aversion are more likely to move in paths which bias toward the bottom-left corner to better avoid the failure states, resulting in paths longer than the shortest possible (e.g., ``Low, High'' team with Intervention Distance: 1). These are the states indicated in Fig.~\ref{fig:maze_3} as either colored white or a light shade of pink. 
Despite the added complexity of the available paths, the teams show strong performance and are frequently near the optimal level of performance. Like the previous case, we should note that deviations of around two are less severe than they appear. In these cases, this is equivalent to a single step off an optimal path and then returning. Further, deviations from an ``optimal'' path account for cases where the agents demonstrate their aversion to the failure states, resulting paths which diverge from the shortest possible. This is most apparent in the cases of agents with a high aversion to risk. While the manager risk aversion is lower, the agent will still choose paths which maintain greater distance from failure states, resulting in paths which require diverging from the shortest possible when risk is not considered. As such, these teams will result in higher $c$ for the manager performance measure. Similar to the previous environment, we see cases with strong alignment of manager and agent risk aversion (e.g., ``None, Low'' and $\delta_I=1$), with corresponding strong team performance at/near optimal. On the other hand, the strong aversion of some agents (e.g., ``Medium, High'' and $\delta_I=1$) results in the longer paths and scores higher than would be possible with agents more willing to take the shorter path near the central failure state. In such a case, there is no manner in which the manager could force the team to take the shorter path, so the resulting score is higher than the minimum/optimal. Again, the manager would only reach this optimal value if the team were able to take a ``riskier'' route to the goal.

As our final case, we utilize a map with several false paths and an overall long trajectory to the goal, which could confuse early exploration. Despite these factors, the results in Table~\ref{tab:hallways} show some of the best scores out of the three scenarios. In most cases, the managed teams achieve the best score, and the only deviations amount on average to less than a single additional intervention or path step.

\section{Conclusion}\label{sec:conclusion}

We presented a manager which provided oversight for teams of RL agents attempting a navigation task in grid environments. These grid environments included states which indicate risky regions of the environment which the agents should avoid. Our manager model was trained and tested with diverse teams of agents. The teams included agents which were trained to navigate the environment with shortest paths which also avoided the risky states. By avoiding risky states, the navigating agents generate paths to the goal state which might deviate from the shortest possible. Managers were given their own notion/measure of risk with which to determine desirable agent delegations. By isolating the notion/model of risk for the agents and managers, we demonstrated how our model could support teams of diverse agents, and we further indicated how our manager could provide oversight while assuming no reliance on knowledge of the agents' measures of desirable behavior. We performed tests with our manager model to compare its performance to that of an optimal model of behavior. Optimality was determined by a minimization of path lengths and occurrences of manager intervention. Our testing demonstrated strong manager performance, with most of our results indicating a manager at or near the optimal result.

\bibliographystyle{IEEEtran}
\bibliography{IEEEabrv,bib_items}

\begin{thebibliography}{10}
\providecommand{\url}[1]{#1}
\csname url@samestyle\endcsname
\providecommand{\newblock}{\relax}
\providecommand{\bibinfo}[2]{#2}
\providecommand{\BIBentrySTDinterwordspacing}{\spaceskip=0pt\relax}
\providecommand{\BIBentryALTinterwordstretchfactor}{4}
\providecommand{\BIBentryALTinterwordspacing}{\spaceskip=\fontdimen2\font plus
\BIBentryALTinterwordstretchfactor\fontdimen3\font minus \fontdimen4\font\relax}
\providecommand{\BIBforeignlanguage}[2]{{%
\expandafter\ifx\csname l@#1\endcsname\relax
\typeout{** WARNING: IEEEtran.bst: No hyphenation pattern has been}%
\typeout{** loaded for the language `#1'. Using the pattern for}%
\typeout{** the default language instead.}%
\else
\language=\csname l@#1\endcsname
\fi
#2}}
\providecommand{\BIBdecl}{\relax}
\BIBdecl

\bibitem{carroll2019utility}
M.~Carroll, R.~Shah, M.~K. Ho, T.~Griffiths, S.~Seshia, P.~Abbeel, and A.~Dragan, ``On the utility of learning about humans for human-ai coordination,'' \emph{Advances in neural information processing systems}, vol.~32, 2019.

\bibitem{chen2018planning}
M.~Chen, S.~Nikolaidis, H.~Soh, D.~Hsu, and S.~Srinivasa, ``Planning with trust for human-robot collaboration,'' in \emph{Proceedings of the 2018 ACM/IEEE international conference on human-robot interaction}, 2018, pp. 307--315.

\bibitem{10.1007/978-3-030-62056-1_44}
Z.~R. Khavas, S.~R. Ahmadzadeh, and P.~Robinette, ``Modeling trust in human-robot interaction: A survey,'' in \emph{Social Robotics}, A.~R. Wagner, D.~Feil-Seifer, K.~S. Haring, S.~Rossi, T.~Williams, H.~He, and S.~Sam~Ge, Eds.\hskip 1em plus 0.5em minus 0.4em\relax Cham: Springer International Publishing, 2020, pp. 529--541.

\bibitem{westby2023collective}
S.~Westby and C.~Riedl, ``Collective intelligence in human-ai teams: A bayesian theory of mind approach,'' in \emph{Proceedings of the AAAI Conference on Artificial Intelligence}, vol.~37, no.~5, 2023, pp. 6119--6127.

\bibitem{wu2021too}
S.~A. Wu, R.~E. Wang, J.~A. Evans, J.~B. Tenenbaum, D.~C. Parkes, and M.~Kleiman-Weiner, ``Too many cooks: Bayesian inference for coordinating multi-agent collaboration,'' \emph{Topics in Cognitive Science}, vol.~13, no.~2, pp. 414--432, 2021.

\bibitem{agudo2024impact}
U.~Agudo, K.~G. Liberal, M.~Arrese, and H.~Matute, ``The impact of ai errors in a human-in-the-loop process,'' \emph{Cognitive Research: Principles and Implications}, vol.~9, no.~1, p.~1, 2024.

\bibitem{cabrera2021discovering}
{\'A}.~A. Cabrera, A.~J. Druck, J.~I. Hong, and A.~Perer, ``Discovering and validating ai errors with crowdsourced failure reports,'' \emph{Proceedings of the ACM on Human-Computer Interaction}, vol.~5, no. CSCW2, pp. 1--22, 2021.

\bibitem{haegler2010no}
K.~Haegler, R.~Zernecke, A.~M. Kleemann, J.~Albrecht, O.~Pollatos, H.~Br{\"u}ckmann, and M.~Wiesmann, ``No fear no risk! human risk behavior is affected by chemosensory anxiety signals,'' \emph{Neuropsychologia}, vol.~48, no.~13, pp. 3901--3908, 2010.

\bibitem{mahmood2022owning}
A.~Mahmood, J.~W. Fung, I.~Won, and C.-M. Huang, ``Owning mistakes sincerely: Strategies for mitigating ai errors,'' in \emph{Proceedings of the 2022 CHI Conference on Human Factors in Computing Systems}, 2022, pp. 1--11.

\bibitem{reason1990human}
J.~Reason, \emph{Human error}.\hskip 1em plus 0.5em minus 0.4em\relax Cambridge university press, 1990.

\bibitem{russell2017human}
S.~Russell, I.~S. Moskowitz, and A.~Raglin, ``Human information interaction, artificial intelligence, and errors,'' \emph{Autonomy and Artificial Intelligence: A Threat or Savior?}, pp. 71--101, 2017.

\bibitem{afanador2019adversarial}
J.~Afanador, M.~Baptista, and N.~Oren, ``An adversarial algorithm for delegation,'' in \emph{Agreement Technologies: 6th International Conference, AT 2018, Bergen, Norway, December 6-7, 2018, Revised Selected Papers 6}.\hskip 1em plus 0.5em minus 0.4em\relax Springer, 2019, pp. 130--145.

\bibitem{chen2023learning}
G.~Chen, X.~Li, C.~Sun, and H.~Wang, ``Learning to make adherence-aware advice,'' \emph{arXiv preprint arXiv:2310.00817}, 2023.

\bibitem{fuchs2024optimizing}
A.~Fuchs, A.~Passarella, and M.~Conti, ``Optimizing delegation in collaborative human-ai hybrid teams,'' 2024.

\bibitem{fuchs2022cognitive}
------, ``A cognitive framework for delegation between error-prone ai and human agents,'' in \emph{2022 IEEE International Conference on Smart Computing (SMARTCOMP)}.\hskip 1em plus 0.5em minus 0.4em\relax IEEE, 2022, pp. 317--322.

\bibitem{fuchs2023compensating}
------, ``Compensating for sensing failures via delegation in human--ai hybrid systems,'' \emph{Sensors}, vol.~23, no.~7, p. 3409, 2023.

\bibitem{fuchs2023optimizing}
------, ``Optimizing delegation between human and ai collaborative agents,'' in \emph{Workshop on Hybrid Human-Machine Learning and Decision Making}, 2023.

\bibitem{aggarwal2014instance}
C.~C. Aggarwal, ``Instance-based learning: A survey.'' \emph{Data classification: algorithms and applications}, vol. 157, 2014.

\bibitem{sutton2018reinforcement}
R.~S. Sutton and A.~G. Barto, \emph{Reinforcement learning: An introduction}.\hskip 1em plus 0.5em minus 0.4em\relax MIT press, 2018.

\bibitem{garcia2020learning}
F.~M. Garcia, C.~Nota, and P.~S. Thomas, ``Learning reusable options for multi-task reinforcement learning,'' \emph{arXiv preprint arXiv:2001.01577}, 2020.

\bibitem{erskine2022developing}
J.~Erskine and C.~Lehnert, ``Developing cooperative policies for multi-stage reinforcement learning tasks,'' \emph{IEEE Robotics and Automation Letters}, vol.~7, no.~3, pp. 6590--6597, 2022.

\bibitem{chakravorty2019option}
J.~Chakravorty, N.~Ward, J.~Roy, M.~Chevalier-Boisvert, S.~Basu, A.~Lupu, and D.~Precup, ``Option-critic in cooperative multi-agent systems,'' \emph{arXiv preprint arXiv:1911.12825}, 2019.

\bibitem{chen2022multi}
J.~Chen, M.~Haliem, T.~Lan, and V.~Aggarwal, ``Multi-agent deep covering option discovery,'' \emph{arXiv preprint arXiv:2210.03269}, 2022.

\bibitem{kurzer2018decentralized}
K.~Kurzer, C.~Zhou, and J.~M. Z{\"o}llner, ``Decentralized cooperative planning for automated vehicles with hierarchical monte carlo tree search,'' in \emph{2018 IEEE intelligent vehicles symposium (IV)}.\hskip 1em plus 0.5em minus 0.4em\relax IEEE, 2018, pp. 529--536.

\bibitem{lau2012coordination}
Q.~P. Lau, M.-L. Lee, and W.~Hsu, ``Coordination guided reinforcement learning.'' in \emph{AAMAS}, 2012, pp. 215--222.

\bibitem{menda2018deep}
K.~Menda, Y.-C. Chen, J.~Grana, J.~W. Bono, B.~D. Tracey, M.~J. Kochenderfer, and D.~Wolpert, ``Deep reinforcement learning for event-driven multi-agent decision processes,'' \emph{IEEE Transactions on Intelligent Transportation Systems}, vol.~20, no.~4, pp. 1259--1268, 2018.

\bibitem{rohanimanesh2002learning}
K.~Rohanimanesh and S.~Mahadevan, ``Learning to take concurrent actions,'' \emph{Advances in neural information processing systems}, vol.~15, 2002.

\bibitem{singh2020hierarchical}
A.~J. Singh, A.~Kumar, and H.~C. Lau, ``Hierarchical multiagent reinforcement learning for maritime traffic management,'' 2020.

\bibitem{yang2022ldsa}
M.~Yang, J.~Zhao, X.~Hu, W.~Zhou, and H.~Li, ``Ldsa: Learning dynamic subtask assignment in cooperative multi-agent reinforcement learning,'' \emph{arXiv preprint arXiv:2205.02561}, 2022.

\bibitem{jacq2022lazy}
A.~Jacq, J.~Ferret, O.~Pietquin, and M.~Geist, ``Lazy-mdps: Towards interpretable rl by learning when to act,'' in \emph{Proceedings of the 21st International Conference on Autonomous Agents and Multiagent Systems}, 2022, pp. 669--677.

\bibitem{balazadeh2020switch}
\BIBentryALTinterwordspacing
V.~B. Meresht, A.~De, A.~Singla, and M.~Gomez{-}Rodriguez, ``Learning to switch between machines and humans,'' \emph{CoRR}, vol. abs/2002.04258, 2020. [Online]. Available: \url{https://arxiv.org/abs/2002.04258}
\BIBentrySTDinterwordspacing

\bibitem{straitouri2021triage}
\BIBentryALTinterwordspacing
E.~Straitouri, A.~Singla, V.~B. Meresht, and M.~Gomez{-}Rodriguez, ``Reinforcement learning under algorithmic triage,'' \emph{CoRR}, vol. abs/2109.11328, 2021. [Online]. Available: \url{https://arxiv.org/abs/2109.11328}
\BIBentrySTDinterwordspacing

\bibitem{richards2016delegate}
D.~Richards and A.~Stedmon, ``To delegate or not to delegate: A review of control frameworks for autonomous cars,'' \emph{Applied ergonomics}, vol.~53, pp. 383--388, 2016.

\bibitem{palmer2020assessing}
S.~Palmer, D.~Richards, G.~Shelton-Rayner, K.~Izzetoglu, and D.~Inch, ``Assessing variable levels of delegated control--a novel measure of trust,'' in \emph{HCI International 2020--Late Breaking Papers: Cognition, Learning and Games: 22nd HCI International Conference, HCII 2020, Copenhagen, Denmark, July 19--24, 2020, Proceedings 22}.\hskip 1em plus 0.5em minus 0.4em\relax Springer, 2020, pp. 202--215.

\bibitem{candrian2022rise}
C.~Candrian and A.~Scherer, ``Rise of the machines: Delegating decisions to autonomous ai,'' \emph{Computers in Human Behavior}, vol. 134, p. 107308, 2022.

\end{thebibliography}

\end{document}